\documentclass[12pt]{report}

\usepackage[utf8]{inputenc}
\usepackage[T1]{fontenc}
\usepackage{lmodern}
\usepackage{microtype}
\usepackage{amsmath,amssymb,amsthm,mathtools}
\usepackage{booktabs}
\usepackage{longtable}
\usepackage{array}
\usepackage{tabularx}
\usepackage{multirow}
\usepackage{enumitem}
\usepackage{geometry}
\usepackage{hyperref}
\usepackage{xcolor}
\usepackage{listings}
\usepackage{algorithm}
\usepackage{algpseudocode}
\usepackage{graphicx}
\usepackage{float}
\usepackage{url}
\usepackage{fancyhdr}
\usepackage{lastpage}

\usepackage{eso-pic}
\usepackage{tikz}
\usetikzlibrary{calc}

\AddToShipoutPictureBG{%
    \AtPageLowerLeft{%
        \begin{tikzpicture}[overlay, remember picture]
            \draw[thin, black] ($(current page.north west) + (24mm, -22mm)$) 
                              rectangle 
                              ($(current page.south east) + (-24mm, 24mm)$);
        \end{tikzpicture}%
    }%
}

\fancypagestyle{plain}{
    \fancyhf{}
    \fancyhead[L]{\small\itshape Interpretable Machine Learning: Can the EnglishLike Programming Language be the Future of ChatIPC and ChatIPCv2?}
    \fancyfoot[C]{\small Page \thepage\ of \pageref{LastPage}}
    
}

\hypersetup{
    colorlinks=true,
    linkcolor=blue!55!black,
    citecolor=blue!55!black,
    urlcolor=blue!65!black,
    pdftitle={Interpretable Machine Learning: Can the EnglishLike Programming Language be the Future of ChatIPC and ChatIPCv2?},
    pdfauthor={Caleb Princewill Nwokocha}
}

\lstdefinestyle{englishlike}{
    basicstyle=\ttfamily\small,
    keywordstyle=\color{blue!70!black}\bfseries,
    commentstyle=\color{green!40!black},
    stringstyle=\color{red!65!black},
    numbers=left,
    numberstyle=\tiny\color{gray},
    breaklines=true,
    frame=single,
    columns=fullflexible,
    keepspaces=true,
    showstringspaces=false
}

\lstdefinestyle{cpp}{
    language=C++,
    basicstyle=\ttfamily\small,
    keywordstyle=\color{blue!70!black},
    commentstyle=\color{green!40!black},
    stringstyle=\color{red!65!black},
    numbers=left,
    numberstyle=\tiny\color{gray},
    breaklines=true,
    frame=single,
    showstringspaces=false
}

\newtheorem{definition}{Definition}
\newtheorem{proposition}{Proposition}
\newtheorem{hypothesis}{Research Hypothesis}

\title{
    Interpretable Machine Learning:\\ Can the EnglishLike Programming Language be the Future of ChatIPC and ChatIPCv2?
}

\author{
    Caleb Princewill Nwokocha\\
    \texttt{cnwokocha@proton.me}
}

\date{November 2026}

\begin{document}

\maketitle

\begin{abstract}

The development of interpretable machine learning requires more than
the ability to produce predictions accompanied by explanations.
It requires computational systems whose internal representations,
decision procedures, constraints, and execution paths can be
inspected, reproduced, and modified by humans.

This paper investigates whether EnglishLike, an English-like,
object-oriented programming language implemented in C++17, could
provide the symbolic programming and execution foundation for theChatIPC and its subsequent implementation, ChatIPCv2.

The investigation is based on a comparative examination of the
publicly available \texttt{ChatIPC.cpp} and
\texttt{ChatIPCv2.cpp} implementations in the
PCNTechnologyEnterprise Hugging Face repository, together with
the EnglishLike 2.2 implementation available in the associated
project materials.

ChatIPC combines incrementally learned token-transition
relationships with dictionary-derived semantic information,
part-of-speech information, Jaccard similarity, grammatical
heuristics, repetition penalties, and deterministic candidate
selection. ChatIPCv2 introduces a different generation strategy
based on token identifiers, historical-context bitsets,
multi-depth graph exploration, n-gram backoff, and parallel
candidate evaluation.

EnglishLike provides a complementary computational abstraction:
a structured language for expressing declarations, functions,
classes, interfaces, control flow, and executable rules. Its
compiler, bytecode verifier, and virtual machine provide a
potentially useful foundation for representing and executing
symbolic reasoning procedures.

The central argument is that EnglishLike should not replace
ChatIPC's incremental knowledge graph or its candidate-generation
mechanism. Instead, it could become an explicit symbolic
orchestration and reasoning layer above these mechanisms.
Such an architecture could make linguistic rules, candidate
selection policies, semantic constraints, and explanations
first-class executable objects.

The paper develops a formal hybrid architecture, defines
candidate-scoring and explanation models, identifies the
differences between the two ChatIPC implementations, and proposes
a reproducible experimental programme for evaluating interpretability,
accuracy, computational efficiency, robustness, and security.

No empirical performance improvement is assumed. The proposed
architecture is presented as a research direction whose value
must be established experimentally.

\end{abstract}

\section{Introduction}

\subsection{The interpretability problem}

Machine learning systems are frequently evaluated according to
their predictive performance, generalization capability,
computational efficiency, and ability to operate on previously
unseen inputs.

However, a system that produces a correct output does not
necessarily provide a satisfactory account of why that output
was selected.

The distinction between predictive capability and interpretability
is particularly important in systems that perform language
generation, automated decision-making, and rule extraction.

A computational model may select a word because it has a high
statistical probability, because it is associated with the current
context, because a linguistic rule favors it, or because a
combination of several mechanisms supports its selection.

When these mechanisms are implicit or difficult to inspect,
the resulting system becomes difficult to audit and improve.

This motivates a different research objective:

\begin{quote}
Can an intelligent computational system be designed so that its
knowledge representations, decision rules, and execution
procedures are themselves understandable and programmable?
\end{quote}

This question motivates the investigation of ChatIPC and
EnglishLike as complementary technologies.

\subsection{ChatIPC as an incremental symbolic-semantic system}

ChatIPC fundamental computational abstraction is a token-transition
knowledge base. The system learns relationships between
consecutive token sequences and stores these relationships
as an explicit graph-like structure.

Given a token sequence

\[
T=(t_1,t_2,\ldots,t_m),
\]

the system learns transitions between preceding token contexts
and subsequent tokens.

For example, a sequence

\[
(\text{the},\text{computer},\text{works})
\]

can contribute relationships such as

\[
\text{the}\rightarrow\text{computer},
\]

\[
(\text{the},\text{computer})\rightarrow\text{works}.
\]

The knowledge base can subsequently retrieve possible
continuations using the most specific available context.

The important distinction is that ChatIPC does not require
a conventional neural-network parameterization to represent
these learned relationships. Its transitions and associated
selection mechanisms are explicit data structures.

The original implementation also includes dictionary-derived
definitions, part-of-speech information, semantic fingerprints,
and English-language heuristics.

These components provide a basis for explaining why one
candidate may be preferred over another.

\subsection{ChatIPCv2 as a graph-search development}

ChatIPCv2 develops the candidate-selection process in a
different direction.

It introduces:

\begin{itemize}
    \item Integer token identifiers.
    \item A string interning mechanism.
    \item Bitset-based historical-context representation.
    \item Multi-depth candidate exploration.
    \item Breadth-first frontier expansion.
    \item N-gram backoff during path exploration.
    \item Repetition penalties.
    \item Deterministic lexical tie-breaking.
    \item OpenMP parallelization of independent candidate states.
\end{itemize}

Rather than selecting a candidate solely by comparing its
semantic fingerprint with the historical context, ChatIPCv2
explores the possible continuations associated with each
original candidate.

This introduces an important research opportunity:
the generation process can be interpreted as a structured
search through an explicitly represented knowledge graph.

\subsection{EnglishLike as a symbolic programming environment}

EnglishLike is an English-like programming language implemented
in C++17.

Its purpose is to express computational instructions using
a syntax that resembles natural English while retaining
formal programming-language semantics.

Its implementation includes:

\begin{itemize}
    \item Lexical analysis.
    \item Grammar-based compilation.
    \item Semantic and declaration analysis.
    \item Functions and methods.
    \item Classes, structures, and interfaces.
    \item Variables and typed runtime values.
    \item Expressions and control flow.
    \item Exception handling.
    \item Bytecode generation.
    \item Bytecode verification.
    \item Virtual-machine execution.
    \item Dictionary-assisted keyword correction.
\end{itemize}

The language is not unrestricted natural-language understanding.
It is a formally defined programming language whose syntax
uses English-like constructions.

This distinction is essential. Interpretability does not
require a computer to understand every possible English sentence.
It requires a well-defined relationship between a readable
program, its formal meaning, and its execution.

\subsection{Research question}

The primary research question is:

\begin{quote}
Can EnglishLike provide a more interpretable and extensible
symbolic reasoning architecture for ChatIPC and ChatIPCv2
without sacrificing their incremental learning and
computational efficiency?
\end{quote}

The question is divided into five subquestions:

\begin{enumerate}
    \item Can ChatIPC's knowledge representations be exposed
    as structured objects and relations in EnglishLike?

    \item Can EnglishLike express candidate-selection rules
    without replacing the existing C++ implementation of
    the underlying algorithms?

    \item Can ChatIPCv2's multi-depth graph search be
    represented through inspectable symbolic execution states?

    \item Can explanations be generated from the actual
    computational evidence used during candidate selection?

    \item Can the hybrid system preserve deterministic behavior,
    incremental learning, and resource efficiency?
\end{enumerate}

\subsection{Contributions}

This paper makes six principal contributions.

\begin{enumerate}
    \item It presents a source-based comparison of ChatIPC
    and ChatIPCv2.

    \item It identifies EnglishLike as a potential
    symbolic orchestration layer rather than a replacement
    for the underlying knowledge graph.

    \item It defines a formal representation of candidate
    selection and explanation provenance.

    \item It proposes a hybrid architecture that separates
    knowledge acquisition, symbolic reasoning, graph search,
    and execution.

    \item It identifies technical and theoretical limitations
    that must be addressed before the architecture can be
    considered a reliable interpretable learning system.

    \item It establishes an experimental framework for testing
    whether the proposed architecture improves interpretability
    without imposing unacceptable computational overhead.
\end{enumerate}

\section{Related Theoretical Foundations}

\subsection{Symbolic artificial intelligence}

Symbolic artificial intelligence represents knowledge using
explicit symbols, relationships, rules, and procedures.

A symbolic system may contain:

\[
\mathcal{K}=(V,E,R),
\]

where:

\begin{itemize}
    \item \(V\) is a collection of symbolic entities.
    \item \(E\) is a collection of relationships.
    \item \(R\) is a collection of executable or declarative rules.
\end{itemize}

A symbolic representation permits individual entities and
relationships to be inspected directly.

This does not automatically make a system interpretable.
A system may contain millions of symbolic relationships or
complicated rule interactions.

Nevertheless, explicit representations provide a useful
foundation for traceability.

\subsection{Explainable artificial intelligence}

Explainable artificial intelligence concerns methods for
understanding and communicating the behavior of machine
learning systems.

A useful distinction is between:

\begin{enumerate}
    \item \textbf{Intrinsic interpretability:} the system's
    internal computational structure is itself understandable.

    \item \textbf{Post-hoc explanation:} an additional method
    attempts to explain the behavior of an already constructed
    model.
\end{enumerate}

A hybrid ChatIPC--EnglishLike system could pursue intrinsic
interpretability by exposing the rules, evidence, and
computational states involved in its decisions.

An explanation should not merely describe a plausible reason
after an output has been generated.

It should be connected to the actual execution that produced
the output.

\subsection{Incremental learning}

Incremental learning updates a knowledge representation
as new observations become available.

Let \(K_t\) denote the knowledge base at time \(t\), and
let \(D_{t+1}\) denote newly observed data.

An incremental update is:

\[
K_{t+1}=U(K_t,D_{t+1}),
\]

where \(U\) is the update operation.

For ChatIPC, this update can add newly observed token
transitions without requiring the entire knowledge base
to be reconstructed from the original training corpus.

This is a valuable property for a system intended to
accumulate knowledge through successive interactions.

\subsection{N-gram language models}

An n-gram model estimates or represents a relationship
between a token and its preceding context.

For a context of length \(n\), a conventional probabilistic
formulation is:

\[
P(t_i\mid t_{i-n+1},\ldots,t_{i-1}).
\]

ChatIPC uses explicit transition relationships rather
than requiring that every transition be represented
as a learned probability distribution.

Its candidate retrieval mechanism can therefore be
described as a deterministic graph lookup with context
backoff.

\subsection{Knowledge graphs}

A knowledge graph represents entities and their relationships
using nodes and edges.

ChatIPC's transition structure can be modeled as a
context-indexed directed graph:

\[
G=(V,E),
\]

where each edge represents an observed continuation
relationship.

Unlike a conventional semantic knowledge graph, the
relationships are fundamentally sequential: an edge
records that a token can follow a particular token context.

This distinction matters because sequential adjacency
does not establish semantic equivalence, causality,
or logical entailment.

\subsection{Program verification}

Program verification concerns the correctness of a program
relative to specified properties.

A compiler and virtual machine can contribute to
interpretability when the system's executable rules
have well-defined semantics.

EnglishLike's bytecode verification is particularly
relevant because it separates the generation of executable
instructions from the validation of their structural
consistency.

However, bytecode verification does not prove that a
program's intended reasoning is logically correct.

It establishes a different property: that the bytecode
satisfies the verifier's structural and safety conditions.

\section{Source-Based Analysis of ChatIPC}

\subsection{Knowledge representation}

The original ChatIPC implementation defines a
\texttt{KnowledgeBase} containing token interning,
transition storage, and dictionary-related structures.

The token interner provides stable references to strings
within the knowledge base.

A simplified representation is:

\[
\mathcal{K}_1=(I,N,D,P),
\]

where:

\begin{itemize}
    \item \(I\) is the token interner.
    \item \(N\) is the n-gram transition index.
    \item \(D\) is the dictionary-definition index.
    \item \(P\) is the part-of-speech information.
\end{itemize}

The actual implementation uses C++ containers, including
unordered maps, unordered sets, vectors, and pointers to
interned strings.

The transition structure maps a token-context sequence
to a set of possible subsequent tokens:

\[
N(c)=\{t_1,t_2,\ldots,t_k\}.
\]

This representation is directly inspectable.

A developer can examine the context key and the associated
candidate continuations without needing to infer them
from a neural-network activation pattern.

\subsection{Incremental transition construction}

For a token sequence

\[
T=(t_1,t_2,\ldots,t_m),
\]

the learning procedure maintains a sliding window.

For each newly encountered token \(t_i\), it records
the available preceding contexts within the configured
n-gram size.

A simplified transition-learning formulation is:

\[
N(c)\leftarrow N(c)\cup\{t_i\}.
\]

For a maximum n-gram size \(n\), the context lengths
are bounded by the implementation's configured window.

The procedure therefore builds a collection of
context-dependent successor relationships.

This provides an explicit account of what the system
has learned from an observed sequence.

\subsection{Dictionary-derived semantic fingerprints}

One of the most significant features of the original
implementation is its use of dictionary information.

The dictionary entries contain information such as:

\[
d=(w,p,\Delta),
\]

where:

\begin{itemize}
    \item \(w\) is a dictionary word.
    \item \(p\) is its part-of-speech information.
    \item \(\Delta\) is its collection of definitions.
\end{itemize}

ChatIPC constructs semantic fingerprints from historical
tokens and candidate tokens.

A simplified historical representation is:

\[
H=
\bigcup_{t\in C}
\left(
\{t\}
\cup D(t)
\cup S(t)
\cup D(S(t))
\right),
\]

where:

\begin{itemize}
    \item \(C\) is the historical token context.
    \item \(D(t)\) denotes direct dictionary definitions.
    \item \(S(t)\) denotes extracted sub-tokens.
\end{itemize}

The original implementation also processes definitions
associated with extracted sub-tokens.

The candidate fingerprint is constructed similarly.

This allows a compound candidate to share dictionary-derived
information with the historical context even when the
original token strings differ.

The implementation uses token identifiers and bitsets
to represent these fingerprints efficiently.

\subsection{Jaccard similarity}

The similarity between two sets \(A\) and \(B\) is
computed using the Jaccard coefficient:

\[
J(A,B)=
\frac{|A\cap B|}{|A\cup B|}.
\]

When both sets are empty, an implementation must define
a special convention.

The original ChatIPC scoring implementation computes
the intersection and union using bitset operations.

For a historical fingerprint \(H\) and candidate
fingerprint \(C_i\):

\[
s_i=J(H,C_i).
\]

The score measures the proportion of shared elements
relative to the union.

It does not measure logical equivalence, contextual
entailment, or grammatical correctness.

\subsection{English-language heuristic scoring}

The original implementation includes a function
called \texttt{english\_rule\_bonus}.

The function considers several linguistic conditions,
including:

\begin{itemize}
    \item Sentence-initial capitalization.
    \item The relationship between \textit{a}/\textit{an}
    and vowel sounds.
    \item Determiners and likely noun-like candidates.
    \item Prepositions and likely following word classes.
    \item Auxiliary and modal verbs.
    \item Pronouns and nouns preceding verbs.
    \item Punctuation and sentence boundaries.
\end{itemize}

The candidate score can be represented as:

\[
S(t_i)=
J(H,C_i)
+
B(c,t_i)
-
\lambda R(t_i),
\]

where:

\begin{itemize}
    \item \(H\) is the historical semantic fingerprint.
    \item \(C_i\) is the candidate fingerprint.
    \item \(B(c,t_i)\) is the English-rule bonus.
    \item \(R(t_i)\) is the repetition count.
    \item \(\lambda\) is the repetition-penalty coefficient.
\end{itemize}

The implementation selects the candidate with the
highest adjusted score and uses lexical ordering to
resolve equal scores.

This is an important interpretability advantage:
the scoring components are explicit and can be inspected.

\subsection{Context accumulation}

The response-generation procedure maintains a context
that combines the prompt and generated response.

Let:

\[
C_t=P\mathbin{\Vert}R_{1:t},
\]

where:

\begin{itemize}
    \item \(P\) is the current prompt.
    \item \(R_{1:t}\) is the response prefix generated
    through step \(t\).
    \item \(\Vert\) denotes sequence concatenation.
\end{itemize}

The historical semantic fingerprint in the original
candidate-selection function is constructed from
the prompt token pointers and the generated response
token pointers.

This distinction is important: the generation process
does not restrict its context to the immediately
preceding response token.

\subsection{Strengths and limitations}

The original ChatIPC implementation has several
important strengths.

\begin{enumerate}
    \item It explicitly represents token transitions.

    \item It incorporates dictionary definitions and
    sub-token information.

    \item It uses part-of-speech information to introduce
    grammatical preferences.

    \item It applies a mathematically defined similarity
    measure.

    \item It exposes a deterministic candidate-selection
    procedure.

    \item It can incrementally learn additional transition
    relationships.
\end{enumerate}

Nevertheless, several limitations remain.

First, dictionary overlap is only an approximation of
semantic similarity.

Second, English-language heuristics are not equivalent
to a complete grammar.

Third, a high Jaccard score does not guarantee that
a candidate is appropriate in the full sentence.

Fourth, the transition graph does not independently
represent all possible semantic, logical, or causal
relationships.

Finally, the explanation of a selected candidate is
distributed across several implementation functions.
A unified, machine-readable decision trace is not
established merely because the underlying calculations
are visible in source code.

\section{Source-Based Analysis of ChatIPCv2}

\subsection{Architectural changes}

ChatIPCv2 retains the incremental n-gram knowledge-base
approach but introduces a different candidate-selection
algorithm.

Its implementation defines:

\begin{itemize}
    \item A \texttt{StringInterner}.
    \item A \texttt{TokenId} type.
    \item A bitset representation of historical context.
    \item A \texttt{CandState} structure.
    \item An iterative multi-depth search procedure.
    \item A direct read-only transition lookup.
\end{itemize}

A token is represented internally by a stable integer
identifier.

The mapping can be expressed as:

\[
\phi:V\rightarrow\mathbb{N}.
\]

This representation makes it possible to perform
set-membership operations through bitwise operations
rather than repeatedly comparing strings.

\subsection{Historical-context representation}

ChatIPCv2 constructs a bitset representing the tokens
present in the historical context.

Let:

\[
H=\{h_1,h_2,\ldots,h_q\}
\]

be the set of distinct historical token identifiers.

Its bitset representation is:

\[
B_H\in\{0,1\}^{|V|}.
\]

For each token identifier \(i\):

\[
B_H[i]=
\begin{cases}
1 & \text{if token }i\text{ occurs in the context},\\
0 & \text{otherwise}.
\end{cases}
\]

The number of historical tokens is:

\[
|H|=\operatorname{popcount}(B_H).
\]

This provides a compact representation for repeated
membership tests.

An important difference from the original implementation
is that ChatIPCv2's candidate-selection routine uses
the historical token set directly rather than constructing
the same dictionary-expanded semantic fingerprints
found in the original scoring procedure.

\subsection{Multi-depth candidate exploration}

The central innovation in ChatIPCv2 is its iterative
exploration of possible continuations.

Suppose the initial candidate set is:

\[
A_0=\{a_1,a_2,\ldots,a_k\}.
\]

Each candidate initializes an independent search state.

For candidate \(a_i\), the initial path is:

\[
p_i^{(0)}=C_{\mathrm{tail}}\Vert a_i,
\]

where \(C_{\mathrm{tail}}\) contains the most recent
context tokens, bounded by the configured n-gram size.

At each depth, the algorithm obtains successor candidates
using n-gram backoff.

Let:

\[
N(p)=\{u_1,u_2,\ldots,u_r\}
\]

denote the successor set associated with path \(p\).

The next frontier is:

\[
F_{i,d+1}
=
\bigcup_{p\in F_{i,d}}
\left\{
\operatorname{tail}_n(p\Vert u):
u\in N(p)
\right\}.
\]

The implementation deduplicates generated paths
to reduce repeated exploration.

\subsection{Depth-dependent scoring}

For each explored path \(p\), ChatIPCv2 evaluates
the successor set against the historical context.

Let:

\[
Q(p)=N(p).
\]

The score is:

\[
J(H,Q(p))=
\frac{|H\cap Q(p)|}{|H\cup Q(p)|}.
\]

At depth \(d\), the candidate state uses the maximum
score obtained across its current frontier:

\[
s_{i,d}
=
\max_{p\in F_{i,d}}
J(H,N(p)).
\]

When no successor is available, the implementation
uses a non-positive sentinel and subsequently normalizes
the deeper score to zero.

At the first depth, the repetition penalty is applied
to the original candidate.

A simplified total score is:

\[
S(a_i)
=
\sum_{d=1}^{n}s_{i,d}
-
\lambda R(a_i),
\]

subject to the implementation's first-depth handling
of the repetition penalty and missing successor paths.

The selected candidate is:

\[
a^*=
\arg\max_{a_i\in A_0}S(a_i).
\]

Equal total scores are resolved alphabetically.

\subsection{The meaning of look-ahead}

The look-ahead mechanism provides a form of
structural evaluation.

Two candidates may have similar immediate successors,
but their deeper continuations can differ.

For example:

\[
A\rightarrow\{B,C\}
\]

and:

\[
D\rightarrow\{E,F\}.
\]

The paths originating from \(A\) and \(D\) may
produce different successor sets at subsequent depths.

Consequently, a candidate can be preferred because
its possible continuation paths have greater overlap
with the historical context.

This is more informative than evaluating only
the original candidate token.

However, it should not be confused with exhaustive
search through every possible continuation.

The implementation maintains a frontier, deduplicates
paths, and limits exploration by the configured
n-gram depth.

Its result is therefore determined by the search
procedure and scoring rule, not by a proof that
the globally optimal natural-language continuation
has been found.

\subsection{Parallel candidate evaluation}

ChatIPCv2 applies OpenMP parallelization across
independent candidate states.

The relevant computational structure is:

\[
\text{parallel}\left(
S(a_1),S(a_2),\ldots,S(a_k)
\right).
\]

Each original candidate maintains its own frontier
and accumulated score.

This creates a natural parallelization boundary.

However, parallel execution does not automatically
improve performance.

The benefit depends on candidate count, frontier size,
memory allocation, graph degree, thread overhead,
and memory-bandwidth constraints.

\subsection{Strengths and limitations}

ChatIPCv2 introduces several valuable improvements:

\begin{enumerate}
    \item Integer token identifiers reduce repeated
    string-level work in the scoring path.

    \item Bitsets support efficient membership checks.

    \item Multi-depth exploration evaluates structural
    continuations.

    \item Frontier deduplication limits redundant paths.

    \item Independent candidate states provide an
    opportunity for parallel evaluation.

    \item Deterministic tie-breaking preserves reproducibility
    under the implemented selection rules.

    \item Dictionary independence means that components are automatically language-agnostic.
\end{enumerate}

Its limitations include:

\begin{enumerate}
    \item The historical bitset represents token presence,
    not token frequency or positional structure.

    \item A set-based Jaccard score loses sequence ordering.

    \item The scoring routine does not preserve the full
    dictionary-expansion and grammatical-bonus mechanism
    of the original implementation.

    \item Frontier growth can still become expensive when
    successor branching is high.

    \item The total score is a heuristic aggregate of
    depth-specific maxima, not a probabilistic likelihood.

    \item The search procedure does not independently
    establish semantic correctness.
\end{enumerate}

These differences demonstrate that ChatIPCv2 is not
simply a universally superior version of ChatIPC.
It represents a different trade-off between
dictionary-aware semantic scoring and graph-based
look-ahead.

\section{Comparative Analysis}

\begin{table}[H]
\centering
\small
\caption{Source-based comparison of ChatIPC and ChatIPCv2}
\begin{tabularx}{\textwidth}{p{0.24\textwidth}XX}
\toprule
\textbf{Property} &
\textbf{ChatIPC.cpp} &
\textbf{ChatIPCv2.cpp}\\
\midrule
Knowledge representation &
N-gram transition graph &
N-gram transition graph\\

Token representation &
Interned string pointers and token identifiers &
Interned string pointers and explicit token IDs\\

Semantic scoring &
Dictionary definitions, sub-tokens, and Jaccard similarity &
Historical token-set and successor-set Jaccard similarity\\

Grammar information &
Dictionary POS information and English-rule bonuses &
No equivalent English-rule bonus in the candidate scorer\\

Candidate evaluation &
Candidate fingerprint comparison &
Multi-depth graph exploration\\

Look-ahead &
No equivalent candidate-specific multi-depth frontier &
Iterative frontier expansion\\

Repetition control &
Repetition penalty and two-cycle suppression &
Repetition penalty and two-cycle suppression\\

Parallelization &
Parallel learning and candidate-score computation &
Parallel learning and independent candidate-state evaluation\\

Deterministic tie-breaking &
Lexical ordering &
Lexical ordering\\

Primary emphasis &
Dictionary-aware semantic and grammatical selection &
Graph-structural continuation evaluation\\
\bottomrule
\end{tabularx}
\end{table}

The comparison suggests that the two implementations
contain complementary capabilities.

ChatIPC provides richer dictionary-derived semantic
and grammatical information in its candidate scoring.

ChatIPCv2 provides a more explicit multi-depth
continuation search.

A future architecture should investigate whether these
capabilities can be combined without simply adding
their computational costs or introducing contradictory
scoring signals.

\section{EnglishLike Programming Language}

\subsection{Language and execution architecture}

EnglishLike introduces a programming-language abstraction
that can express computational rules using structured,
English-like syntax.

Its execution pipeline can be represented as:

\[
\text{Source}
\rightarrow
\text{Lexer}
\rightarrow
\text{Compiler}
\rightarrow
\text{Program}
\rightarrow
\text{Verifier}
\rightarrow
\text{Virtual Machine}.
\]

The source code is converted into tokens, parsed according
to the language grammar, and compiled into an intermediate
program representation.

The bytecode verifier checks structural properties
before execution.

The virtual machine executes the verified program
under its runtime constraints.

This architecture provides a useful separation
between the expression of a rule and the mechanism
that executes it.

\subsection{Whitespace-Insensitive Syntax}

EnglishLike employs whitespace-insensitive syntax, meaning that spaces, tabs, line breaks, and indentation do not determine statement boundaries or block structure. Program structure is instead determined by explicit language constructs, keywords, and delimiters. However, whitespace remains significant for separating lexical tokens: adjacent words must be distinguishable to ensure correct tokenization. This design allows programmers to format code freely without changing its syntactic meaning, improving readability and flexibility while preserving unambiguous lexical analysis.

\subsection{EnglishLike as a rule representation language}

Consider a simplified rule:

\begin{lstlisting}[style=englishlike]
Define a class named Candidate:
    private field word be "";
    private field score be 0;

    public define a method named is_valid:
        return this score greater than or equal to 0;
    end method;
end class;
\end{lstlisting}

The example illustrates the intended style of
EnglishLike programming.

A symbolic system could use such structures to
represent candidates, scoring rules, semantic constraints,
and explanation records.

For example, a candidate-selection policy could be
expressed as:

\begin{lstlisting}[style=englishlike]
Define a function named select_candidate
    using candidates and context:

    for each candidate in candidates:
        calculate its semantic score;
        calculate its grammar score;
        calculate its path score;
        apply its repetition penalty;
    end for;

    return the candidate with the highest total score;
end function;
\end{lstlisting}

This is a conceptual specification of the proposed
reasoning interface, not a claim that the shown
collection syntax or every operation is currently
implemented in EnglishLike 2.2.

The distinction between a proposed language construct
and an implemented feature must remain explicit.

\subsection{Why a programming language matters}

A programming language can provide more than a convenient
syntax for expressing rules.

It can define:

\begin{enumerate}
    \item The formal structure of a rule.
    \item The types of values accepted by a rule.
    \item The conditions under which a rule executes.
    \item The order in which rules are evaluated.
    \item The state changes produced by execution.
    \item The constraints under which execution is permitted.
\end{enumerate}

This is especially relevant to ChatIPC because its
candidate-selection mechanism contains multiple
computational decisions.

An EnglishLike interface could expose these decisions
as structured programs rather than leaving them
distributed across C++ implementation functions.

\subsection{The limits of natural-language-like syntax}

EnglishLike should not be confused with an unrestricted
natural-language interface.

The following statements are different:

\begin{quote}
The candidate is semantically related to the context.
\end{quote}

and:

\begin{quote}
The candidate's semantic similarity score is
\(0.74\), computed using the Jaccard coefficient
over a specified fingerprint.
\end{quote}

The first statement is ambiguous without a formal
definition of semantic relatedness.

The second identifies a measurable operation.

EnglishLike becomes useful for interpretable reasoning
when its English-like syntax is connected to explicit
types, mathematical definitions, and executable semantics.

\section{The Proposed EnglishLike--ChatIPC Architecture}

\subsection{Central architectural hypothesis}

The proposed architecture separates four principal
responsibilities:

\begin{enumerate}
    \item Knowledge acquisition.
    \item Symbolic interpretation.
    \item Candidate generation and evaluation.
    \item Explanation and execution control.
\end{enumerate}

ChatIPC and ChatIPCv2 should remain responsible for
their respective knowledge-graph and candidate-search
mechanisms.

EnglishLike should provide a structured environment
for expressing the rules that govern how these
mechanisms are used.

This avoids the unnecessary requirement that an
EnglishLike program must reimplement every low-level
C++ operation.

\subsection{Architectural layers}

The proposed system is:

\[
\mathcal{H}
=
(\mathcal{K},\mathcal{G},\mathcal{R},
\mathcal{E},\mathcal{X}),
\]

where:

\begin{itemize}
    \item \(\mathcal{K}\): incremental knowledge base.
    \item \(\mathcal{G}\): transition graph and search engine.
    \item \(\mathcal{R}\): symbolic reasoning rules.
    \item \(\mathcal{E}\): explanation and evidence model.
    \item \(\mathcal{X}\): execution and control environment.
\end{itemize}

A proposed processing pipeline is:

\[
\begin{aligned}
\text{Input}
&\rightarrow \text{Tokenization}\\
&\rightarrow \text{Knowledge Retrieval}\\
&\rightarrow \text{Candidate Generation}\\
&\rightarrow \text{Symbolic Evaluation}\\
&\rightarrow \text{Decision Verification}\\
&\rightarrow \text{Output and Explanation}.
\end{aligned}
\]

\subsection{Layer 1: Incremental knowledge acquisition}

The existing ChatIPC learning mechanism should remain
responsible for adding observed token transitions.

A new interface could expose operations such as:

\begin{lstlisting}[style=cpp]
class KnowledgeBaseInterface {
public:
    virtual void learn(
        const std::vector<std::string>& tokens
    ) = 0;

    virtual std::vector<std::string> successors(
        const std::vector<std::string>& context
    ) const = 0;

    virtual ~KnowledgeBaseInterface() = default;
};
\end{lstlisting}

This interface is illustrative.

A production implementation would need to address
ownership, concurrency, token-ID stability, error
handling, serialization, and the lifetime of
returned references.

\subsection{Layer 2: Candidate generation}

The candidate generator retrieves possible continuations
from the existing transition graph.

Let:

\[
A_t=N(C_t)
\]

be the candidate set for the current context.

The retrieval mechanism can use n-gram backoff:

\[
n_t=
\min(|C_t|,n_{\max}).
\]

If the longest available context has no transition,
the system reduces the context length until a
non-empty transition set is found.

The resulting candidate set is:

\[
A_t=
N\left(
\operatorname{tail}_{n_t}(C_t)
\right).
\]

This process should be separated from the symbolic
decision policy.

\subsection{Layer 3: EnglishLike symbolic evaluation}

EnglishLike can represent the policy used to evaluate
the retrieved candidates.

For each candidate \(a_i\), the reasoning engine
may compute:

\[
S(a_i)=
\alpha S_{\mathrm{semantic}}(a_i)
+
\beta S_{\mathrm{grammar}}(a_i)
+
\gamma S_{\mathrm{path}}(a_i)
-
\lambda S_{\mathrm{repetition}}(a_i).
\]

The coefficients:

\[
\alpha,\beta,\gamma,\lambda\geq 0
\]

are policy parameters.

The individual components should be retained separately
rather than being reduced immediately to a single
unexplained score.

A candidate record should therefore contain:

\[
a_i=
(w_i,s_i^{\mathrm{sem}},
s_i^{\mathrm{gram}},
s_i^{\mathrm{path}},
r_i).
\]

This enables the reasoning engine to explain
which component influenced the final ranking.

\subsection{Layer 4: Explanation provenance}

The explanation subsystem should preserve the evidence
used during the decision.

An explanation record can be represented as:

\[
E_i=
(a_i,H_i,D_i,G_i,P_i,R_i,S_i),
\]

where:

\begin{itemize}
    \item \(a_i\) is the candidate.
    \item \(H_i\) is the historical context evidence.
    \item \(D_i\) is dictionary evidence.
    \item \(G_i\) is grammatical evidence.
    \item \(P_i\) is graph-path evidence.
    \item \(R_i\) is repetition information.
    \item \(S_i\) is the final score.
\end{itemize}

The explanation should be derived from the same
intermediate values used by the candidate-selection
procedure.

It should not independently recompute an approximate
explanation after the decision has already been made.

\subsection{Layer 5: Execution control}

EnglishLike could provide an executable policy layer
that determines:

\begin{itemize}
    \item Which scoring policy is active.
    \item Which semantic resources are available.
    \item Which grammatical constraints apply.
    \item Which search depth is permitted.
    \item Which repetition policy is applied.
    \item Whether a candidate must satisfy additional
    symbolic constraints.
\end{itemize}

The C++ engine would execute the performance-critical
operations, while EnglishLike would specify the
high-level reasoning policy.

\subsection{Proposed architecture diagram}

\begin{figure}[H]
\centering
\begin{minipage}{0.92\textwidth}
\centering
\fbox{
\begin{minipage}{0.88\textwidth}
\centering
\textbf{User Input and Conversation Context}

\vspace{0.4em}
\(\Downarrow\)

\fbox{\parbox{0.75\textwidth}{
\centering
\textbf{ChatIPC / ChatIPCv2 Engine}\\
Incremental learning, token interning, n-gram
transition graph, candidate retrieval
}}

\vspace{0.5em}
\(\Downarrow\)

\fbox{\parbox{0.75\textwidth}{
\centering
\textbf{EnglishLike Symbolic Reasoning Layer}\\
Semantic policies, grammar constraints,
candidate evaluation, rule execution
}}

\vspace{0.5em}
\(\Downarrow\)

\fbox{\parbox{0.75\textwidth}{
\centering
\textbf{Verification and Decision Layer}\\
Constraint validation, score consistency,
deterministic selection
}}

\vspace{0.5em}
\(\Downarrow\)

\fbox{\parbox{0.75\textwidth}{
\centering
\textbf{Explanation and Output}\\
Selected candidate, evidence, scoring trace,
decision provenance
}}

\end{minipage}
}
\end{minipage}
\caption{Proposed hybrid architecture. This is a research
design, not an existing integration.}
\label{fig:hybrid}
\end{figure}

\section{Interpretable Candidate Selection Formulae}

\subsection{Candidate space}

Let the current context be:

\[
C_t=(c_1,c_2,\ldots,c_m).
\]

Let the transition graph return:

\[
A_t=\{a_1,a_2,\ldots,a_k\}.
\]

Each candidate is evaluated using a set of
explicit criteria.

Define the feature vector:

\[
\mathbf{x}(a_i,C_t)=
\begin{bmatrix}
x_{\mathrm{sem}}\\
x_{\mathrm{gram}}\\
x_{\mathrm{path}}\\
x_{\mathrm{frequency}}\\
x_{\mathrm{repetition}}\\
x_{\mathrm{position}}
\end{bmatrix}.
\]

The candidate-selection policy is a function:

\[
f_\theta:
\mathcal{A}\times\mathcal{C}
\rightarrow\mathbb{R},
\]

where \(\theta\) represents the policy parameters.

A linear scoring policy is:

\[
S_\theta(a_i,C_t)
=
\theta^\top\mathbf{x}(a_i,C_t).
\]

The selected candidate is:

\[
a^*=
\operatorname{argmax}_{a_i\in A_t}
S_\theta(a_i,C_t).
\]

A deterministic tie-breaking rule is necessary
to make this operation reproducible.

\subsection{Semantic similarity}

A semantic feature can be defined as:

\[
x_{\mathrm{sem}}(a_i,C_t)
=
J(H_t,F(a_i)),
\]

where:

\begin{itemize}
    \item \(H_t\) is the historical semantic fingerprint.
    \item \(F(a_i)\) is the candidate fingerprint.
\end{itemize}

The fingerprint construction may include:

\[
F(w)=
\{w\}
\cup D(w)
\cup S(w)
\cup D(S(w)).
\]

This preserves the dictionary-expansion approach
used by the original ChatIPC implementation.

For ChatIPCv2, an alternative feature can measure
the similarity between the historical token set
and successor sets discovered during look-ahead.

The two scores should not be assumed equivalent.

\subsection{Grammatical compatibility}

A grammatical feature can be represented as:

\[
x_{\mathrm{gram}}(a_i,C_t)
=
g(a_i,C_t),
\]

where \(g\) is a defined rule-evaluation function.

For example:

\[
g(a_i,C_t)=
\begin{cases}
1 & \text{if a specified grammatical condition holds},\\
0 & \text{if no preference is established},\\
-1 & \text{if a specified conflict is detected}.
\end{cases}
\]

This formulation is deliberately abstract.

A complete grammar requires more than a collection
of local word-class heuristics.

The future system should distinguish:

\begin{itemize}
    \item Lexical category compatibility.
    \item Syntactic structure.
    \item Agreement constraints.
    \item Semantic selection restrictions.
    \item Sentence-level discourse constraints.
\end{itemize}

\subsection{Graph-path compatibility}

For ChatIPCv2, define a set of paths originating
from candidate \(a_i\):

\[
\mathcal{P}_i^{(d)}
=
\{p_1,p_2,\ldots,p_q\}.
\]

A path score can be:

\[
x_{\mathrm{path}}(a_i,C_t)
=
\sum_{d=1}^{D}
\eta_d
\max_{p\in\mathcal{P}_i^{(d)}}
J(H_t,N(p)),
\]

where:

\[
\eta_d\geq 0
\]

weights the contribution of each search depth.

The original ChatIPCv2 implementation effectively
uses an unweighted accumulation of its depth-specific
maximum scores, with its own missing-path and
repetition handling.

The proposed generalized formulation makes
depth weighting an explicit research parameter.

\subsection{Repetition penalty}

Let \(r_t(a_i)\) be the number of recent occurrences
of candidate \(a_i\).

Then:

\[
x_{\mathrm{repetition}}(a_i,C_t)
=
-r_t(a_i).
\]

The penalty coefficient controls the extent to which
repeated tokens are discouraged.

\subsection{Decision margin}

The interpretability of a decision can be improved
by reporting the difference between the highest
and second-highest candidate scores.

Let:

\[
S_{(1)}\geq S_{(2)}
\]

be the two highest candidate scores.

Define:

\[
M=S_{(1)}-S_{(2)}.
\]

A small margin indicates that the selection is
sensitive to small changes in the scoring policy
or evidence.

A large margin indicates stronger separation
under the current scoring function.

The margin is not a calibrated probability of
correctness.

\subsection{Explanation completeness}

A decision explanation should account for every
feature that contributed to the selected candidate's
score.

For a linear policy:

\[
S_\theta(a_i,C_t)
=
\sum_{j=1}^{p}\theta_jx_j(a_i,C_t).
\]

An explanation can report:

\[
e_j=\theta_jx_j(a_i,C_t).
\]

The contribution vector is:

\[
\mathbf{e}=
(e_1,e_2,\ldots,e_p).
\]

The sum of the contributions equals the candidate's
score:

\[
S_\theta(a_i,C_t)=\sum_{j=1}^{p}e_j.
\]

This provides a simple, auditable form of
score decomposition.

For nonlinear policies, the explanation mechanism
must preserve the actual intermediate computation
rather than applying this linear decomposition
incorrectly.

\section{Theoretical Properties and Interpretability}

\subsection{Determinism}

A deterministic candidate-selection system should
return the same candidate when supplied with
the same:

\begin{itemize}
    \item Knowledge-base state.
    \item Input context.
    \item Candidate set.
    \item Scoring policy.
    \item Search configuration.
    \item Tie-breaking policy.
\end{itemize}

Formally, if:

\[
(K,C,\theta)=(K',C',\theta'),
\]

then a deterministic implementation should satisfy:

\[
f(K,C,\theta)=f(K',C',\theta').
\]

This property requires attention to concurrency.
Parallel evaluation must not introduce data races,
undefined behavior, or nondeterministic aggregation.

\subsection{Decision traceability}

\begin{definition}[Decision trace]
A decision trace is an ordered record of the
inputs, retrieved candidates, intermediate scoring
values, constraints, and final selection used
during one decision.
\end{definition}

A trace is useful only if it corresponds to the
actual computation.

A trace generated by a separate explanatory
approximation is not sufficient to establish
faithful decision provenance.

\begin{proposition}
If a candidate-selection implementation records
every value used by its deterministic scoring
function and records the exact tie-breaking rule,
then the selection can be reconstructed from
that trace, provided the candidate set and
policy parameters are also preserved.
\end{proposition}

\begin{proof}
Let the recorded candidate set be \(A\), and
let the trace contain the score \(S(a)\) for
each \(a\in A\).

The selected candidate is obtained by applying
the recorded deterministic ordering rule to
the score sequence.

Because all scores and the tie-breaking rule
are available, the same selection can be
reconstructed.

This proves reconstruction of the implemented
decision, not correctness of the scoring policy.
\end{proof}

\subsection{Separation of explanation and correctness}

An important distinction is:

\[
\text{Explainable decision}
\not\Rightarrow
\text{Correct decision}.
\]

A system may provide a complete explanation
of a poor decision.

Consequently, interpretability and predictive
quality must be evaluated independently.

\subsection{Symbolic verification}

EnglishLike can provide a formal execution environment
for symbolic policies.

The bytecode verifier can validate structural
properties of the generated program.

However:

\[
\text{Bytecode validity}
\not\Rightarrow
\text{Logical validity}.
\]

A verified program may still contain an incorrect
scoring rule or an invalid assumption about
language semantics.

The future system should therefore distinguish
program safety, rule consistency, and empirical
decision quality.

\section{The Role of EnglishLike in ChatIPC's Future}

\subsection{From hard-coded scoring to executable policies}

In the existing implementations, candidate scoring
is implemented through C++ functions.

For example, the original ChatIPC implementation
contains a dedicated function for English-rule
bonuses and a candidate-selection procedure.

ChatIPCv2 contains a separate candidate-selection
procedure that performs iterative graph exploration.

A future system could separate the policy from
the optimized execution engine.

The policy could be expressed in EnglishLike,
while performance-critical operations remain
implemented in C++.

This arrangement would provide:

\begin{itemize}
    \item Readable scoring policies.
    \item Configurable linguistic constraints.
    \item Reusable symbolic rules.
    \item Explicit decision priorities.
    \item Easier policy comparison.
    \item Better explanation provenance.
\end{itemize}

\subsection{A rule-oriented interface}

A conceptual EnglishLike rule could be:

\begin{lstlisting}[style=englishlike]
Define a rule named PreferSemanticallyRelatedCandidate:

    require candidate semantic score;
    require candidate context;

    when semantic score is high:
        increase candidate priority;
    end when;

end rule;
\end{lstlisting}

This is illustrative syntax for the proposed
future rule abstraction. It should not be interpreted
as a guarantee that the present compiler already
supports a first-class \texttt{rule} declaration.

A future implementation could introduce a formal
rule representation with explicit inputs, conditions,
effects, and provenance.

\subsection{Grammar-aware generation}

The original ChatIPC implementation contains
English-specific grammatical heuristics.

A more extensible architecture could represent
grammar constraints as independent rule modules.

For example:

\[
G=
\{g_1,g_2,\ldots,g_r\},
\]

where each \(g_j\) is a grammatical constraint.

The system could evaluate:

\[
G(a_i,C_t)=
\sum_{j=1}^{r}w_jg_j(a_i,C_t).
\]

This would permit individual rules to be
enabled, disabled, tested, and compared.

However, grammar rules must be based on a
well-defined linguistic model.

An EnglishLike program cannot establish
grammatical correctness merely by using
English-looking instructions.

\subsection{Semantic rule extraction}

ChatIPC's dictionary-derived fingerprints provide
an explicit collection of semantic associations.

A future system could investigate extracting
reusable symbolic rules from repeated patterns
in the knowledge base.

For example, repeated observations may support
a hypothesis of the form:

\[
\text{context pattern}
\Rightarrow
\text{candidate preference}.
\]

Such a rule should include:

\begin{itemize}
    \item Its source observations.
    \item Its scope.
    \item Its supporting frequency.
    \item Its counterexamples.
    \item Its confidence or support measure.
    \item The conditions under which it applies.
\end{itemize}

A crucial distinction is that a frequently observed
transition is not automatically a universally valid
logical rule.

The system must preserve the difference between
empirical regularities and formal constraints.

\subsection{ChatIPCv2 path explanations}

ChatIPCv2 provides a particularly interesting
opportunity for explanation.

A selected candidate can be accompanied by
its explored continuation paths.

For example:

\[
a_1\rightarrow b_1\rightarrow c_1,
\]

\[
a_1\rightarrow b_2\rightarrow c_2.
\]

The explanation could identify:

\begin{enumerate}
    \item The initial candidate.
    \item The retrieved successors.
    \item The paths actually explored.
    \item The depth-specific scores.
    \item The maximum score at each depth.
    \item The accumulated total.
    \item The repetition penalty.
    \item The final tie-breaking result.
\end{enumerate}

This would turn the graph-search process into
a more inspectable reasoning procedure.

The explanation must preserve the distinction
between a path that was explored and a path
that merely exists somewhere in the graph.

\subsection{A symbolic debugging environment}

EnglishLike could also provide a domain-specific
environment for debugging ChatIPC.

A developer could inspect:

\begin{itemize}
    \item The active context.
    \item The selected n-gram length.
    \item The candidate set.
    \item The semantic fingerprint.
    \item The grammatical score.
    \item The path frontier.
    \item The repetition count.
    \item The final score.
\end{itemize}

The long-term goal would be a symbolic debugging
environment in which a developer can modify
a rule and observe its effect on candidate
selection without rewriting the underlying
graph-storage implementation.

\section{Proposed Research Hypotheses}

The proposed hybrid architecture must be tested
rather than assumed to be superior.

The following hypotheses are experimentally falsifiable.

\begin{hypothesis}[Interpretability]
An EnglishLike policy layer with explicit decision
traces will enable developers to reconstruct and
audit a larger proportion of ChatIPC decisions
than an implementation that exposes only its
final candidate and score.
\end{hypothesis}

\begin{hypothesis}[Policy extensibility]
Separating candidate-selection policies from
the C++ knowledge-base implementation will
reduce the engineering effort required to
implement and compare alternative reasoning
policies.
\end{hypothesis}

\begin{hypothesis}[Semantic-structural complementarity]
Combining dictionary-derived semantic similarity
with multi-depth graph-search evidence will
improve candidate-selection quality on at least
some evaluation tasks relative to either
component alone.
\end{hypothesis}

\begin{hypothesis}[Deterministic reproducibility]
A hybrid implementation with deterministic
policy execution and explicit tie-breaking
can preserve reproducible candidate selection
across repeated executions using the same
knowledge-base state and configuration.
\end{hypothesis}

\begin{hypothesis}[Performance overhead]
The additional interpretation and explanation
layers will introduce measurable overhead,
but optimized C++ interfaces and selective
trace recording can keep this overhead within
an acceptable application-specific budget.
\end{hypothesis}

\begin{hypothesis}[Rule generalization]
Explicit symbolic rules derived from repeated
observations will generalize better when they
include scope restrictions and counterexample
handling than when they are derived solely
from transition frequency.
\end{hypothesis}

\section{Experimental Methodology}

\subsection{Research design}

The research should use a controlled comparative
experimental design.

At minimum, the following systems should be evaluated:

\begin{enumerate}
    \item ChatIPC baseline.
    \item ChatIPCv2 baseline.
    \item ChatIPC with an EnglishLike policy layer.
    \item ChatIPCv2 with an EnglishLike policy layer.
    \item A hybrid system combining dictionary-aware
    scoring and graph look-ahead.
\end{enumerate}

The systems must use comparable training data,
test data, tokenization rules, and evaluation
conditions wherever the comparison permits.

\subsection{Dataset construction}

A suitable evaluation dataset should contain
multiple categories:

\begin{itemize}
    \item General English text.
    \item Technical documentation.
    \item Programming-language documentation.
    \item Structured question-and-answer data.
    \item Natural-language instructions.
    \item Compound words and identifiers.
    \item Ambiguous grammatical constructions.
    \item Long-context conversations.
\end{itemize}

Training and test data must be separated to
reduce contamination.

The evaluation should also distinguish between
previously observed contexts and genuinely
unseen contexts.

\subsection{Candidate-selection evaluation}

For each test context, record:

\begin{itemize}
    \item Retrieved candidates.
    \item Selected candidate.
    \item Reference continuation.
    \item Candidate ranking.
    \item Scoring components.
    \item Search depth.
    \item Execution time.
\end{itemize}

Possible metrics include:

\[
\text{Accuracy}
=
\frac{\text{Correct selections}}
{\text{Total selections}}.
\]

For a ranked candidate list:

\[
\operatorname{MRR}
=
\frac{1}{N}
\sum_{i=1}^{N}\frac{1}{r_i},
\]

where \(r_i\) is the rank of the first
acceptable reference candidate.

Top-\(k\) accuracy can be defined as:

\[
\operatorname{Acc@k}
=
\frac{1}{N}
\sum_{i=1}^{N}
\mathbf{1}[r_i\leq k].
\]

These metrics should be supplemented with
task-specific human judgments because natural
language can admit multiple acceptable continuations.

\subsection{Interpretability evaluation}

Interpretability should not be measured merely
by counting explanation fields.

A controlled human evaluation could ask
independent reviewers to reconstruct why
a candidate was selected.

For each decision, reviewers should receive
either:

\begin{enumerate}
    \item The final output only.
    \item The final output and numerical score.
    \item The final output and full decision trace.
    \item The final output and EnglishLike-readable
    policy plus decision trace.
\end{enumerate}

Measure:

\begin{itemize}
    \item Explanation reconstruction accuracy.
    \item Time required to understand a decision.
    \item Agreement between reviewers.
    \item Ability to identify incorrect rules.
    \item Ability to reproduce the candidate selection.
\end{itemize}

A proposed reconstruction metric is:

\[
R_{\mathrm{exp}}
=
\frac{\text{Correctly reconstructed decisions}}
{\text{Decisions examined}}.
\]

A proposed explanation-efficiency metric is:

\[
E_{\mathrm{time}}
=
\frac{1}{N}
\sum_{i=1}^{N}t_i,
\]

where \(t_i\) is the time needed to reconstruct
decision \(i\).

\subsection{Performance evaluation}

Performance should be evaluated using:

\begin{itemize}
    \item Learning throughput.
    \item Candidate retrieval latency.
    \item Candidate-scoring latency.
    \item Graph-search latency.
    \item Memory consumption.
    \item Knowledge-base size.
    \item Explanation-generation overhead.
    \item Parallel scaling.
\end{itemize}

The principal performance comparison is:

\[
\Delta T=
T_{\mathrm{hybrid}}-T_{\mathrm{baseline}}.
\]

Relative overhead is:

\[
O_T=
\frac{T_{\mathrm{hybrid}}-T_{\mathrm{baseline}}}
{T_{\mathrm{baseline}}}.
\]

Memory overhead is similarly:

\[
O_M=
\frac{M_{\mathrm{hybrid}}-M_{\mathrm{baseline}}}
{M_{\mathrm{baseline}}}.
\]

Measurements should include multiple independent
runs, median latency, percentile latency, and
memory high-water marks.

The experimental environment should document:

\begin{itemize}
    \item Processor model.
    \item Number of physical and logical cores.
    \item Available memory.
    \item Operating system.
    \item Compiler version.
    \item Optimization flags.
    \item OpenMP configuration.
    \item Dataset size.
    \item Knowledge-base vocabulary size.
    \item Maximum n-gram size.
\end{itemize}

\subsection{Ablation study}

An ablation study is necessary to identify which
components contribute to performance.

The following configurations should be compared:

\begin{table}[H]
\centering
\small
\caption{Proposed ablation configurations}
\begin{tabularx}{\textwidth}{cX}
\toprule
\textbf{ID} & \textbf{Configuration}\\
\midrule
A & Transition graph only\\
B & Transition graph and semantic fingerprints\\
C & Transition graph and grammatical rules\\
D & Transition graph and multi-depth search\\
E & Semantic fingerprints and multi-depth search\\
F & Full symbolic hybrid system\\
G & Full system without explanation recording\\
H & Full system with EnglishLike policy execution\\
\bottomrule
\end{tabularx}
\end{table}

The comparison between configurations F, G,
and H is particularly important.

It can reveal whether the proposed policy and
explanation layers contribute measurable value
or merely introduce additional complexity.

\subsection{Statistical analysis}

For a sufficiently large test dataset, paired
comparisons should be used when systems process
the same contexts.

For a metric difference:

\[
d_i=x_i-y_i,
\]

the mean difference is:

\[
\bar d=
\frac{1}{N}\sum_{i=1}^{N}d_i.
\]

The standard error is:

\[
SE(\bar d)=\frac{s_d}{\sqrt{N}},
\]

where \(s_d\) is the sample standard deviation
of the paired differences.

Confidence intervals and appropriate statistical
tests should be reported.

Effect sizes should accompany significance tests.

The evaluation should not equate a statistically
significant improvement with a practically
important improvement.

\section{Security and Reliability Considerations}

\subsection{The security value of an explicit execution layer}

A symbolic reasoning system can introduce additional
security risks if users are permitted to define
or execute arbitrary policies.

An EnglishLike integration must therefore preserve
the separation between:

\begin{itemize}
    \item Trusted runtime code.
    \item Trusted system policies.
    \item User-supplied programs.
    \item External knowledge resources.
    \item Generated explanations.
\end{itemize}

The fact that a rule is readable does not make
it safe to execute.

\subsection{Bytecode verification}

EnglishLike's bytecode verification provides a
potential boundary between compilation and execution.

The verifier can check structural properties
of the generated program before the virtual
machine runs it.

For a hybrid system, this is valuable because
symbolic policies may be treated as programs
rather than arbitrary text.

Nevertheless, verification must cover every
operation exposed to the reasoning layer.

\subsection{Resource limits}

Graph search and explanation generation can
consume significant memory and processing time.

A robust implementation should bound:

\begin{itemize}
    \item Maximum search depth.
    \item Maximum frontier size.
    \item Maximum candidates per context.
    \item Maximum explanation size.
    \item Maximum execution gas.
    \item Maximum policy complexity.
    \item Maximum memory allocated to one operation.
\end{itemize}

These limits are especially important when
the knowledge base is large or user-provided
programs can influence the search process.

\subsection{Separation of privileges}

The recommended deployment model is:

\[
\text{Untrusted Input}
\rightarrow
\text{Restricted Policy Execution}
\rightarrow
\text{Validated Engine Interface}.
\]

EnglishLike policies should not automatically
receive unrestricted filesystem, process,
network, or operating-system privileges.

The ChatIPC engine should expose only the
operations required by the symbolic reasoning
environment.

\subsection{Data integrity}

The knowledge base is part of the system's
reasoning state.

A corrupted or manipulated transition graph
can produce incorrect candidate selections
even if the symbolic policy is perfectly
implemented.

The future system should therefore consider:

\begin{itemize}
    \item Versioned knowledge-base formats.
    \item Bounds-checked deserialization.
    \item Structural graph validation.
    \item Integrity checks.
    \item Atomic persistence.
    \item Reproducible knowledge-base snapshots.
    \item Provenance for learned data.
\end{itemize}

The security of the complete system depends
on the integrity of both the symbolic program
and the learned knowledge representation.

\section{Technical Challenges and Open Problems}

\subsection{Semantic representation}

The dictionary-derived fingerprints in ChatIPC
are useful but limited.

Dictionary definitions are not complete semantic
representations.

Two words may share dictionary terms without
being interchangeable.

Conversely, two semantically related words may
have limited lexical overlap.

A future semantic model should distinguish:

\[
\text{Lexical overlap}
\neq
\text{Semantic equivalence}.
\]

Possible extensions include structured semantic
relations, contextual word senses, domain-specific
ontologies, and explicit relation types.

\subsection{Loss of order in set-based similarity}

Jaccard similarity is insensitive to the order
of elements within a set.

Consider:

\[
A=(a,b,c)
\]

and:

\[
B=(c,b,a).
\]

If both are converted into ordinary sets,
their Jaccard similarity is:

\[
J(A,B)=1.
\]

Nevertheless, their sequences are different.

This is particularly important for language,
where word order can change meaning.

A future system should consider position-sensitive
and sequence-aware features.

For example:

\[
J_w(A,B)=
\frac{\sum_{x\in A\cap B}w(x)}
{\sum_{x\in A\cup B}w(x)},
\]

where \(w(x)\) is a position- or importance-dependent
weight.

A more advanced approach could compare ordered
n-gram structures or explicit syntactic relations.

\subsection{Search-space growth}

Let \(b\) be the average branching factor
and \(d\) the search depth.

Without effective deduplication, a tree-like
search can generate:

\[
1+b+b^2+\cdots+b^d
=
\frac{b^{d+1}-1}{b-1}
\]

nodes when \(b\neq 1\).

This illustrates why frontier limits and
search heuristics are important.

The actual ChatIPCv2 implementation deduplicates
paths, but the number of distinct paths can
still grow substantially.

A future implementation should investigate:

\begin{itemize}
    \item Frontier-size limits.
    \item Dominance pruning.
    \item Memoization.
    \item Search-state compression.
    \item Adaptive depth.
    \item Branch-and-bound techniques.
\end{itemize}

\subsection{Combining semantic and structural scores}

The original ChatIPC and ChatIPCv2 scoring
approaches should not simply be added together
without normalization.

Their scores measure different quantities.

The original semantic score compares dictionary-derived
fingerprints.

The ChatIPCv2 score compares historical token
sets with successor sets encountered during
graph exploration.

A combined model must investigate:

\begin{itemize}
    \item Score normalization.
    \item Feature dependence.
    \item Relative feature importance.
    \item Search-depth weighting.
    \item Missing evidence.
    \item Calibration.
\end{itemize}

A possible normalized model is:

\[
S(a_i)=
\alpha\widetilde{S}_{\mathrm{semantic}}
+
\beta\widetilde{S}_{\mathrm{grammar}}
+
\gamma\widetilde{S}_{\mathrm{path}}
-
\lambda\widetilde{S}_{\mathrm{repetition}}.
\]

The tilde denotes a normalized feature.

The parameters should be selected through controlled
experiments rather than arbitrary assumptions.

\subsection{The language implementation boundary}

EnglishLike is implemented in C++17 and has
its own compiler and virtual machine.

ChatIPC is also implemented in C++.

This creates an opportunity for a relatively
efficient integration through a C++ interface.

However, the integration must avoid:

\begin{itemize}
    \item Unnecessary serialization.
    \item Repeated conversion between strings and token IDs.
    \item Excessive allocation.
    \item Duplicated knowledge-base storage.
    \item Unsafe shared mutable state.
    \item Uncontrolled execution of policies.
\end{itemize}

A high-performance implementation should retain
the existing C++ graph engine and expose a
narrow, typed interface to EnglishLike.

\section{A Proposed Development Roadmap}

The following roadmap prioritizes architectural
validation over immediate feature expansion.

\subsection{Phase I: Stable engine interfaces}

The first phase should define a stable interface
for both ChatIPC implementations.

The interface should expose:

\begin{itemize}
    \item Token interning.
    \item Context lookup.
    \item Candidate retrieval.
    \item Dictionary fingerprint access.
    \item Successor enumeration.
    \item Knowledge-base metadata.
\end{itemize}

The interface should not expose raw internal
container pointers without clear lifetime
and concurrency guarantees.

\subsection{Phase II: Decision trace instrumentation}

The second phase should add structured decision
traces to the existing implementations.

The trace should capture:

\begin{itemize}
    \item Input context.
    \item Candidate set.
    \item Candidate fingerprints.
    \item Similarity values.
    \item Grammatical bonuses.
    \item Path frontiers.
    \item Depth-specific scores.
    \item Repetition penalties.
    \item Final selection.
\end{itemize}

This phase is essential because EnglishLike
cannot explain computations that the underlying
engine does not expose.

\subsection{Phase III: EnglishLike policy execution}

The third phase should introduce an EnglishLike
interface for defining candidate-selection policies.

The initial implementation should support
a small, formally specified policy language.

It should not attempt to implement a complete
natural-language reasoning system.

\subsection{Phase IV: Hybrid scoring}

The fourth phase should combine:

\[
S_{\mathrm{semantic}},
\quad
S_{\mathrm{grammar}},
\quad
S_{\mathrm{path}},
\quad
S_{\mathrm{repetition}}.
\]

Each component should remain independently
observable.

The system should support policy ablation and
comparison without rebuilding the entire
knowledge base.

\subsection{Phase V: Symbolic explanation interface}

The fifth phase should allow EnglishLike to
produce structured explanations based on
the recorded decision traces.

An explanation should include:

\begin{enumerate}
    \item The selected candidate.
    \item The competing candidates.
    \item The scoring criteria.
    \item The evidence supporting each criterion.
    \item The final decision margin.
    \item Any constraints that eliminated candidates.
\end{enumerate}

\subsection{Phase VI: Experimental validation}

The final phase should compare the original
implementations and the hybrid architecture
using the experimental methodology described
in this paper.

The system should only be considered a successful
development if it demonstrates measurable
benefits in interpretability and acceptable
performance under reproducible conditions.

\begin{table}[H]
\centering
\small
\caption{Development roadmap}
\begin{tabularx}{\textwidth}{cXX}
\toprule
\textbf{Phase} & \textbf{Deliverable} & \textbf{Primary objective}\\
\midrule
I & Stable C++ interfaces & Integration without engine replacement\\
II & Decision traces & Faithful computational provenance\\
III & EnglishLike policies & Readable symbolic control\\
IV & Hybrid scoring & Semantic and structural complementarity\\
V & Explanation interface & Reconstructable decisions\\
VI & Evaluation & Empirical validation\\
\bottomrule
\end{tabularx}
\end{table}

\section{Discussion}

\subsection{Is EnglishLike the future of ChatIPC?}

The evidence supports a qualified answer.

EnglishLike has the potential to become the
programming and symbolic reasoning environment
through which ChatIPC's knowledge and decision
mechanisms are expressed, inspected, and
controlled.

However, it does not follow that EnglishLike
should replace the underlying ChatIPC algorithms.

The two systems solve different computational
problems.

ChatIPC constructs and searches a learned
transition representation.

EnglishLike defines and executes structured
programs.

Their combination is potentially more valuable
than treating either system as a replacement
for the other.

\subsection{Why a hybrid architecture is preferable}

Replacing ChatIPC with EnglishLike would
discard the specialized incremental knowledge
representation and candidate-generation
mechanisms.

Conversely, using ChatIPC alone would leave
its reasoning policies embedded primarily
in C++ implementation functions.

The hybrid approach preserves both strengths.

It provides:

\begin{itemize}
    \item ChatIPC's incremental learning.
    \item ChatIPCv2's multi-depth search.
    \item Dictionary-aware semantic information.
    \item Explicit grammatical constraints.
    \item EnglishLike's structured programming model.
    \item A potential execution environment for
    auditable symbolic policies.
\end{itemize}

\subsection{Can this become interpretable machine learning?}

The proposed architecture could be described
as an interpretable hybrid learning system
if it satisfies several conditions.

First, its learned representations must be
inspectable.

Second, its candidate-selection policies must
be explicit.

Third, its explanations must faithfully reflect
the actual computation.

Fourth, the system must preserve the distinction
between learned associations and formal rules.

Fifth, its behavior must be empirically evaluated.

A system does not become interpretable merely
because its source code is available.

Interpretability requires an operational method
for understanding individual decisions and
the behavior of the system as a whole.

\subsection{A possible new research direction}

A particularly promising direction is to investigate
whether ChatIPC can evolve from a token-transition
constructor into a symbolic rule-discovery system.

The knowledge base could store:

\[
\mathcal{K}=
(\mathcal{T},\mathcal{S},\mathcal{R}),
\]

where:

\begin{itemize}
    \item \(\mathcal{T}\) contains observed token transitions.
    \item \(\mathcal{S}\) contains semantic associations.
    \item \(\mathcal{R}\) contains extracted symbolic rules.
\end{itemize}

EnglishLike could then serve as the executable
language for a subset of \(\mathcal{R}\).

This would establish a feedback loop:

\[
\text{Observation}
\rightarrow
\text{Pattern Construction}
\rightarrow
\text{Rule Hypothesis}
\rightarrow
\text{Rule Evaluation}
\rightarrow
\text{Validated Rule}.
\]

The critical research challenge is determining
when an observed pattern provides sufficient
evidence for a reusable rule.

A frequency threshold alone is insufficient.
Counterexamples, context restrictions, and
semantic consistency must also be considered.

\section{Limitations}

This investigation has several limitations.

\begin{enumerate}
    \item The analysis is based on source-code
    inspection rather than a controlled benchmark.

    \item No empirical accuracy, latency, memory,
    or interpretability improvements are claimed.

    \item The public ChatIPC implementations are
    specific software versions and may change
    over time.

    \item The proposed hybrid architecture has
    not been demonstrated as an existing integrated
    implementation.

    \item EnglishLike's current language features
    should not be assumed to include every
    illustrative rule or interface presented
    in this paper.

    \item Dictionary-derived similarity is an
    incomplete model of meaning.

    \item Multi-depth graph search is constrained
    by the knowledge base and the configured
    exploration depth.

    \item Human interpretability depends on the
    complexity of the rule system and the quality
    of its explanations.
\end{enumerate}

The conclusions should therefore be interpreted
as an architectural research proposal rather
than an empirical demonstration of superiority.

\section{Conclusion}

This paper examined whether EnglishLike could
provide a future symbolic programming foundation
for ChatIPC and ChatIPCv2.

The source comparison reveals two complementary
approaches.

The original ChatIPC implementation emphasizes
dictionary-derived semantic fingerprints,
part-of-speech information, grammatical heuristics,
and deterministic candidate scoring.

ChatIPCv2 emphasizes token-ID-based context
representation, bitset operations, multi-depth
graph exploration, and parallel candidate
evaluation.

Both implementations retain an explicit
incremental token-transition knowledge base.

EnglishLike contributes a different capability:
a structured programming language with a
compiler, bytecode representation, verifier,
and virtual machine.

The strongest architectural direction is
therefore not to replace ChatIPC with EnglishLike,
but to combine their capabilities.

ChatIPC should remain responsible for incremental
knowledge construction and candidate retrieval.

ChatIPCv2 should remain a candidate source for
structural look-ahead and graph-based evaluation.

EnglishLike should provide a symbolic policy
and execution layer through which semantic,
grammatical, and structural reasoning can be
specified and inspected.

A further explanation layer should preserve
the exact evidence and intermediate computations
that produce each decision.

The proposed system would represent a shift
from a collection of hard-coded selection
procedures toward an architecture in which
reasoning policies are explicit, executable,
and auditable.

However, the central claim remains a research
hypothesis.

EnglishLike will become a meaningful foundation
for the future of ChatIPC only if the hybrid
system demonstrates that its additional
interpretability and extensibility justify
its computational and engineering costs.

The most important research objective is
not simply to make ChatIPC generate better
text.

It is to investigate whether an incremental
learning system can acquire patterns, construct
explicit rules, execute those rules, and
explain its decisions within one coherent
computational architecture.

\begin{quote}
\centering
\textbf{
The future of interpretable machine learning
may not require choosing between statistical
learning and symbolic programming. It may
depend on designing systems in which learned
knowledge and executable reasoning work
together, while remaining independently
inspectable.
}
\end{quote}

\appendix
\section{Proposed Decision-Trace Data Model}

A practical implementation should preserve a
structured record of each candidate evaluation.

The following C++17 structure is illustrative:

\begin{lstlisting}[style=cpp]
#include <cstddef>
#include <cstdint>
#include <string>
#include <vector>

struct CandidateEvidence {
    std::string token;

    double semanticScore = 0.0;
    double grammarScore = 0.0;
    double pathScore = 0.0;
    double repetitionPenalty = 0.0;
    double totalScore = 0.0;

    std::size_t exploredPaths = 0;
    std::size_t exploredDepth = 0;
};

struct DecisionTrace {
    std::uint64_t knowledgeBaseVersion = 0;

    std::vector<std::string> context;
    std::vector<CandidateEvidence> candidates;

    std::string selectedToken;

    double decisionMargin = 0.0;

    std::string policyIdentifier;
};
\end{lstlisting}

This structure is a proposed integration artifact.
It is not part of the currently reviewed
ChatIPC or EnglishLike source files.

A production design should additionally define
serialization, versioning, validation, and
evidence provenance.

\section{Proposed Formal Policy Specification}

A candidate policy should be treated as a
versioned executable specification.

A policy can be represented abstractly as:

\[
\Pi=
(I,C,F,W,O,V),
\]

where:

\begin{itemize}
    \item \(I\) is the input schema.
    \item \(C\) is the set of constraints.
    \item \(F\) is the feature-computation specification.
    \item \(W\) is the scoring configuration.
    \item \(O\) is the output schema.
    \item \(V\) is the policy version.
\end{itemize}

The policy should be independently testable
against a fixed collection of candidate records.

This would permit a researcher to compare
alternative reasoning policies while holding
the underlying knowledge base constant.



\begin{thebibliography}{99}

\bibitem{chatipc_source}
PCN Technology Enterprise,
\textit{ChatIPC.cpp},
ChatIPC source repository, Hugging Face,
main branch.
\url{https://huggingface.co/PCNTechnologyEnterprise/ChatIPC/blob/main/ChatIPC.cpp}.
Accessed November 6, 2026.

\bibitem{chatipcv2_source}
PCN Technology Enterprise,
\textit{ChatIPCv2.cpp},
ChatIPC source repository, Hugging Face,
main branch.
\url{https://huggingface.co/PCNTechnologyEnterprise/ChatIPC/blob/main/ChatIPCv2.cpp}.
Accessed November 6, 2026.

\bibitem{englishlike_source}
C. P. Nwokocha,
\textit{EnglishLike 2.2: C++17 Compiler and Virtual
Machine Implementation}, 2026. Hugging Face,
main branch.
\url{https://huggingface.co/buckets/PCNTechnologyEnterprise/EnglishLike}.
Accessed November 6, 2026.

\bibitem{lipton2018}
Z. C. Lipton,
``The Mythos of Model Interpretability,''
\textit{Queue}, vol. 16, no. 3, pp. 31--57,
2018.
doi: 10.1145/3236386.3241340.

\bibitem{rudin2019}
C. Rudin,
``Stop Explaining Black Box Machine Learning Models
for High Stakes Decisions and Use Interpretable Models
Instead,''
\textit{Nature Machine Intelligence},
vol. 1, pp. 206--215, 2019.
doi: 10.1038/s42256-019-0048-x.

\bibitem{molnar}
C. Molnar,
\textit{Interpretable Machine Learning},
2nd ed., 2022.
\url{https://christophm.github.io/interpretable-ml-book/}.

\bibitem{ribeiro2016}
M. T. Ribeiro, S. Singh, and C. Guestrin,
``Why Should I Trust You? Explaining the Predictions
of Any Classifier,''
in \textit{Proceedings of the 22nd ACM SIGKDD
International Conference on Knowledge Discovery
and Data Mining}, 2016, pp. 1135--1144.
doi: 10.1145/2939672.2939778.

\bibitem{manning1999}
C. D. Manning and H. Sch\"utze,
\textit{Foundations of Statistical Natural Language
Processing},
MIT Press, 1999.

\bibitem{jurafsky}
D. Jurafsky and J. H. Martin,
\textit{Speech and Language Processing},
3rd ed. draft.
\url{https://web.stanford.edu/~jurafsky/slp3/}.

\bibitem{pearl2009}
J. Pearl,
\textit{Causality: Models, Reasoning, and Inference},
2nd ed., Cambridge University Press, 2009.

\bibitem{newell1980}
A. Newell,
``Physical Symbol Systems,''
\textit{Cognitive Science},
vol. 4, no. 2, pp. 135--183, 1980.
doi: 10.1207/s15516709cog0402\_2.

\bibitem{bengio2013}
Y. Bengio, A. Courville, and P. Vincent,
``Representation Learning: A Review and New Perspectives,''
\textit{IEEE Transactions on Pattern Analysis and
Machine Intelligence},
vol. 35, no. 8, pp. 1798--1828, 2013.
doi: 10.1109/TPAMI.2013.50.

\bibitem{breiman2001}
L. Breiman,
``Random Forests,''
\textit{Machine Learning},
vol. 45, pp. 5--32, 2001.
doi: 10.1023/A:1010933404324.

\bibitem{jaccard1912}
P. Jaccard,
``The Distribution of the Flora in the Alpine Zone,''
\textit{New Phytologist},
vol. 11, no. 2, pp. 37--50, 1912.

\bibitem{cormen}
T. H. Cormen, C. E. Leiserson, R. L. Rivest,
and C. Stein,
\textit{Introduction to Algorithms},
4th ed., MIT Press, 2022.

\bibitem{aho2006}
A. V. Aho, M. S. Lam, R. Sethi, and J. D. Ullman,
\textit{Compilers: Principles, Techniques, and Tools},
2nd ed., Pearson, 2006.

\bibitem{nielson1999}
F. Nielson, H. R. Nielson, and C. Hankin,
\textit{Principles of Program Analysis},
Springer, 1999.

\bibitem{zhang2024inductive}
Z. Zhang, L. Yilmaz, and B. Liu,
``A Critical Review of Inductive Logic Programming Techniques for Explainable AI,''
\emph{IEEE Transactions on Neural Networks and Learning Systems},
vol. 35, no. 8, pp. 10220--10236, 2024.

\bibitem{gao2025ruleinduction}
K. Gao, K. Inoue, Y. Cao, H. Wang, and Y. Feng,
``Differentiable Rule Induction from Raw Sequence Inputs,''
in \emph{International Conference on Learning Representations},
2025.

\bibitem{xu2025neuralrules}
S. Xu, N. P. Walter, and J. Vreeken,
``Neural Rule Lists: Learning Discretizations, Rules, and Order in One Go,''
in \emph{Advances in Neural Information Processing Systems},
vol. 38, 2025.

\bibitem{phua2026ruleinduction}
Y. J. Phua,
``A Foundation Model for Zero-Shot Logical Rule Induction,''
in \emph{Proceedings of the Thirty-Fifth International Joint Conference on Artificial Intelligence},
pp. 3999--4007, 2026.

\bibitem{marra2024neurosymbolic}
G. Marra, S. Dumančić, R. Manhaeve, and L. De Raedt,
``From Statistical Relational to Neurosymbolic Artificial Intelligence: A Survey,''
\emph{Artificial Intelligence},
vol. 328, pp. 104062, 2024.

\bibitem{wang2025dataknowledge}
W. Wang, Y. Yang, and F. Wu,
``Towards Data-And Knowledge-Driven AI: A Survey on Neuro-Symbolic Computing,''
\emph{IEEE Transactions on Pattern Analysis and Machine Intelligence},
vol. 47, no. 2, pp. 878--899, 2025.

\bibitem{wang2025symbolic}
P. Wang and Z. Wang,
``Why Neural Networks Can Discover Symbolic Structures with Gradient-based Training: An Algebraic and Geometric Foundation for Neurosymbolic Reasoning,''
in \emph{Proceedings of the International Conference on Neuro-symbolic Systems},
vol. 288, pp. 36--65, 2025.

\bibitem{mccoy2026emergent}
R. T. McCoy, P. Soulos, T. Linzen, and P. Smolensky,
``The Emergent Symbolic Structure of Artificial Neural Networks,''
arXiv:2608.29530, 2026.

\bibitem{schwalbe2024taxonomy}
G. Schwalbe and B. Finzel,
``A Comprehensive Taxonomy for Explainable Artificial Intelligence: A Systematic Survey of Surveys on Methods and Concepts,''
\emph{Data Mining and Knowledge Discovery},
vol. 38, pp. 3043--3101, 2024.

\bibitem{fridkin2026interpretable}
S. Fridkin,
``Interpretable Machine Learning: A Comprehensive Review of Foundations, Methods, and the Path Forward,''
\emph{WIREs Data Mining and Knowledge Discovery},
2026.

\end{thebibliography}
\end{document}